\documentclass[letterpaper, 10 pt, conference]{ieeeconf}  %

\IEEEoverridecommandlockouts                              %

\usepackage{graphics} %
\usepackage{epsfig} %
\usepackage{mathptmx} %
\usepackage{times} %
\usepackage{amsmath} %
\usepackage{amssymb}  %
\usepackage{comment} %
\usepackage{hyperref}

\title{\LARGE \bf
The body image of social robots*
}

\author{Bing Li$^{1}$, Oumayma Ajjaji$^{1}$, Robin Gigandet$^{1}$ and Tatjana
  Nazir$^{1}$%
  \thanks{*This project has received funding from the MEL (M\'etropole
    europ\'eenne de Lille) and the I-SITE ULNE (Universit\'e Lille
    Nord-Europe). Grant awarded to TAN (n$^\circ$:
    R-Talent-20-006-Nazir).}%
  \thanks{$^{1}$The authors are affiliated with SCALAB, UMR CNRS 9193,
    Universit\'e de Lille, Lille, France. Address correspondence to Bing Li
    ({\tt\small bing.li@univ-lille.fr}) or Tatjana Nazir ({\tt\small
      tatjana.nazir@univ-lille.fr})}%
}

\begin{document}

\maketitle
\thispagestyle{empty}
\pagestyle{empty}

\begin{abstract}

  The rapid development of social robots has challenged robotics and cognitive
  sciences to understand humans' perception of the appearance of robots.  In
  this study, robot-associated words spontaneously generated by humans were
  analyzed to semantically reveal the body image of 30 robots that have been
  developed over the past decades.  The analyses took advantage of word affect
  scales and embedding vectors, and provided a series of evidence for links
  between human perception and body image.  It was found that the valence and
  dominance of the body image reflected humans' attitude towards the general
  concept of robots; that the user bases and usages of the robots were among the
  primary factors influencing humans' impressions towards individual robots; and
  that there was a relationship between the robots' affects and semantic
  distances to the word ``person''.  According to the results, building body
  image for robots was an effective paradigm to investigate which features were
  appreciated by people and what influenced people's feelings towards robots.
  
\end{abstract}

\section{INTRODUCTION}

The present study aims to understand how humans perceive robots and how they
form semantic representations of the robots from purely visual
characteristics. The term ``body image'' will be used to refer to the semantic
understanding a person derives from viewing (images of) robots.

Neuroscientific research has shown that visual input to the brain is transformed
into semantic representations through a default path
\cite{doerig_semantic_2022}. Visual traits such as strokes, shapes, and colors
are processed in the occipital lobe and then transmitted to the frontal and
temporal lobes through two pathways for semantic interpretation of the object
\cite{kar_fast_2021}. Importantly, significant traits such as faces and word
forms are processed in dedicated regions close to occipital regions
\cite{hannagan_emergence_2021,tsao_faces_2003}, before other semantic
information is processed in the temporal lobe
\cite{patterson_where_2007,bracci_dissociations_2016,tong_distributed_2022}. Recently,
a third pathway has been proposed that is specialized in processing social cues
\cite{pitcher_evidence_2021}. Therefore, the body image of a robot encompasses
not only the visual appearance of the robot but also its semantic meaning and
social significance.

Studies in semantics provided models for how concepts are organized in the
brain's semantic system. In psycholinguistics, a commonly used model is the word
embedding model \cite{bhatia_associative_2017}. This model calculates the
likelihood of words appearing next to each other in large, naturally occurring
texts typically found on the internet using a single-layer neural
network. Research has shown that the model accurately reflects real-world
semantic information \cite{grand_semantic_2022}. Therefore, incorporating a
person's understanding of a robot's functions into such a semantic framework
could potentially improve the model of a robot's body image. Instead of focusing
on the visual appearance of robots that are often difficult to quantify, we thus
propose to deduce a robot's body image by analyzing words that are spontaneously
associate with the robot.

With the goal of understanding human's mental representation of social robots,
we propose to examine the characteristic features of robots from a semantic
perspective. We will use a free word-association task to collect basic
impressions about various robots and model these data within a sematic space
using word vectors. We will also quantify the affect dimensions of the words
using ``valence'', ``arousal'' and ``dominance'' measurements from an affect
lexicon \cite{mohammad_obtaining_2018}. Finally, we will determine how the
word-vector that characterize a given robot relate to the word ``person'' within
the semantic space, to assess how closely participants associate the robot with
human environment. Taken together, the study demonstrates the importance of
representation in studying human-robot interaction.

\begin{table}[t!]
  \caption{The attitude questionnaire used to measure participants' attitudes
    towards the general concept of robots. It preceded the free association
    task.}
  \label{tab:jhBl}
  \centering
\begin{tabular}{|l|l|}
\hline
INSTRUCTION & Please select the option that best\\
 & describes your attitude at the moment:\\
\hline
STATEMENTS & I want a robot to assist me \ldots{}\\
 & \hspace{1mm} (1)  \ldots{} at home\\
 & \hspace{1mm} (2)  \ldots{} at school\\
 & \hspace{1mm} (3)  \ldots{} in dangerous locations\\
 & \hspace{1mm} (4)  \ldots{} in factories\\
 & \hspace{1mm} (5)  \ldots{} in hospitals\\
 & \hspace{1mm} (6)  \ldots{} in hotels\\
 & \hspace{1mm} (7)  \ldots{} in museums\\
 & \hspace{1mm} (8)  \ldots{} in offices\\
 & \hspace{1mm} (9)  \ldots{} in police stations\\
 & \hspace{1mm} (10) \ldots{} in public transportations\\
 & \hspace{1mm} (11) \ldots{} in shopping centers\\
 & \hspace{1mm} (12) \ldots{} in sports facilities\\
\hline
RESPONSES & \(\square\) Strongly disagree\\
 & \(\square\) Disagree\\
 & \(\square\) Neither agree or disagree\\
 & \(\square\) Agree\\
 & \(\square\) Strongly agree\\
\hline
\end{tabular}

\end{table}

\begin{table}[hb!]
  \caption{The frequency of the top 30 words in the free association task, based
    on all the 30 participants and 30 robots.}
  \label{tab:nVHv}
  \centering
\begin{tabular}{|r|l|r|}
\hline
 & Word & Frequency\\
\hline
1 & toy & 42\\
2 & cute & 39\\
3 & friendly & 33\\
4 & small & 30\\
5 & robot & 29\\
6 & helpful & 28\\
7 & scary & 25\\
8 & future & 23\\
9 & dog & 22\\
10 & creepy & 17\\
11 & fun & 16\\
12 & human & 16\\
13 & technology & 16\\
14 & child & 14\\
15 & cool & 14\\
16 & weird & 14\\
17 & animal & 12\\
18 & simple & 12\\
19 & happy & 11\\
20 & smart & 11\\
21 & strong & 11\\
22 & artificial intelligence & 10\\
23 & assistant & 10\\
24 & automatic & 10\\
25 & color & 10\\
26 & helper & 10\\
27 & modern & 10\\
28 & useful & 10\\
29 & wheels & 10\\
30 & automated & 9\\
\hline
\end{tabular}

\end{table}

\section{METHODS}

\subsection {The robots and human participants}

We selected 30 robots from the robots available on the market or as prototypes
in the past decades. The robots are picked to make a wide range of diverse
features, e.g. whether the robot has face, legs, arms, or not resemble common
animal- or human-shapes at all. The robots have been included into the study
are: Buddy, Eilik, Musio, Pico, Sima, Aibo, Ameca, Aquanaut1, Asimo, Astro,
Atlas, Emiew, Hexa, Hitchbot, iCub, Jibo, Lovot, Minicheetah, Moxie, Nao,
Neubie, Nicobo, Optimus, Parky, Pepper, Pyxel, Sawyer, Spotmini, Talos, Vector.

It's worth mentioning that robot is an ambiguous concept, and sometimes it's
hard to decide which ones should be included. A typical case is drones,
e.g. quadrotor aircraft. Although some people would regard them as robots
(e.g. drones are listed as robots at \url{https://robots.ieee.org/robots/}),
some others would consider it simply as aircraft. On the other side, the purpose
of the present study is to investigate the robot body image as a basis for
human-robot interaction, which is not a typical purpose of drones. Therefore
they were excluded from the present study.

The experiment was conducted on the Prolific
platform \url{https://app.prolific.co} and built with psychJS
\cite{de_jspsych_2015}. Before publishing the experiment task, we set
pre-screening filters to only include English native speaker between 18 and 30
year old to maintain the homogeneity of the age span, within which people are
very likely to grow up with internet, and more open to new
technologies. %

Before the tasks, participants were informed that they can quit the task at any
moment during the task and will be compensated for the time spent on the
task. The task averagely took 14 minutes per participant. Compensation is about
€7.15/hour. Thirty participants were recruited into the experiment.

\subsection {The robot attitude scale}

The experiment was made up of two sessions, one session measures the attitudes
to robots, and the other session builds body images of the robots. To measure
the attitudes, we designed a questionnaires. And to build the body images of
robots, we used a free association task
\cite{nelson_free_2000,deyne_better_2013}. The participants were required to
first finish the questionnaires, and then the free association task.

The robot attitude questionnaire is a Likert scale presenting statements about
whether the subject would like to be assist by robots in different scenarios, as
shown in Tab. \ref {tab:jhBl}. The questionnaire is modified from a scale used
by another study examining the differences across cultures
\cite{lee_culturally_2014}. Participants chose the option that best describes
the participants' attitude on a 5-point Likert scale: from 1 - ``Strongly
disagree'' to 5 - ``Strongly agree''. The responses were attributed with values
from 0 to 4, representing a spectrum from negative to positive attitude. The
overall attitude within a participant was averaged over all statements. This
questionnaire came before the free association task, hence \textit{robot} was
not supposed to represent any specific robot model but just the general concept
of robots.

\subsection {Free association}

After the questionnaires, we asked the participants to type 6 words that come
into their minds for each of 10 images presented serially. Each image was
presented at the top of the screen, followed by 6 text boxes. The participants
needed to type in the words sequentially from the 1st to the 6th text box, and
click on the ``Finished'' button to proceed to the next image. They were able to
press ``Enter'' key to quickly jump to the next word after typing a word. For
each image, all the 6 words were mandatory and no text box can be skipped.  The
``Auto fill-in'' functionality of our psychJS program was disabled.

In order to avoid the confounding of participants intra-correlation, i.e. a
participant acts similar across the robots observed, we used a randomized robots
selection procedure. The randomization was described in another study
\cite{hollis_scoring_2018}. With this procedure, every participant will see a
different subset of 10 robot.

We listed the top 30 most frequent words in Tab. \ref {tab:nVHv}.

\subsection {Word vectors and affect measurements}
\begin{figure*}[t!]
  \centering
  \includegraphics[width=.7\linewidth] {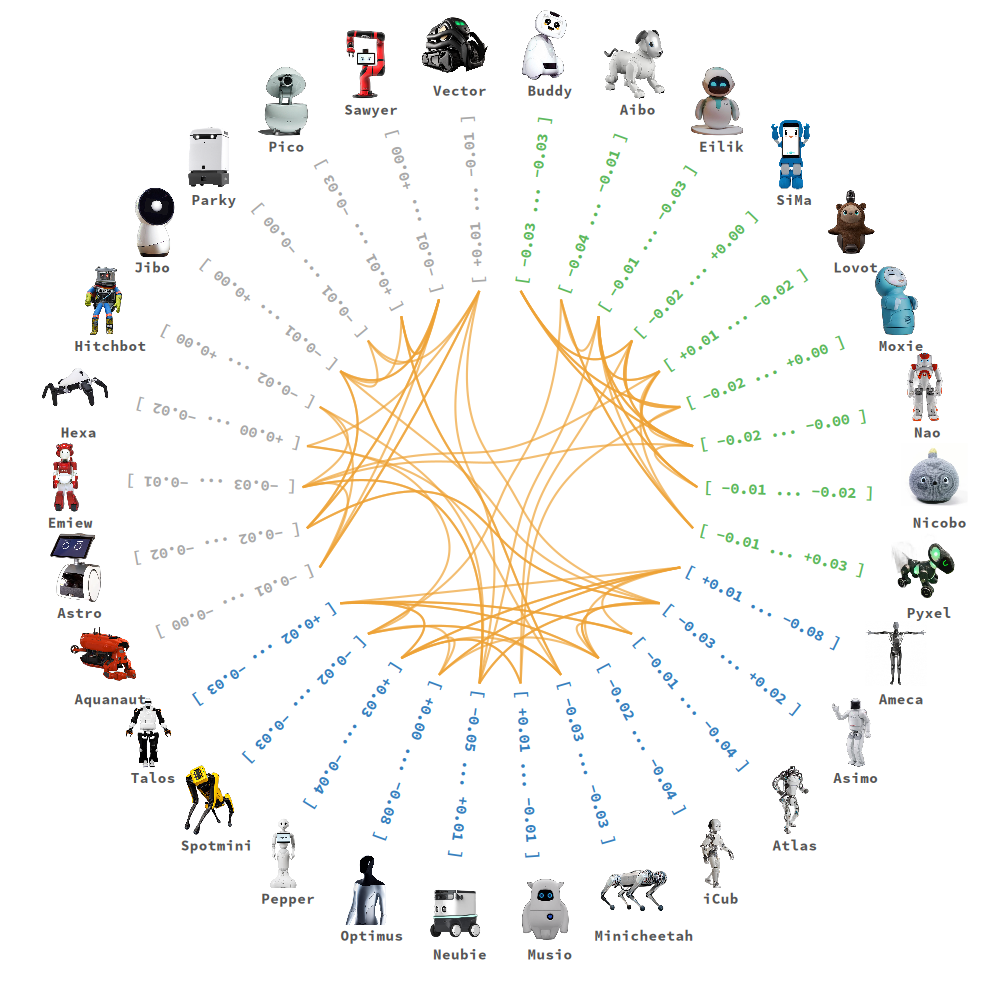}
  \caption{Clustering of the robots based on their cosine similarities.  The
    outermost circle is the robots in the present study. The intermediate circle
    indicates the vector abstraction of the robots, colored with the three
    clusters identified by hierarchical clustering of the cosine similarities
    between robots. And the innermost network is comprised of the connections
    between each robot and its 3 closest neighbors.}
  \label{fig:cCUq}
\end{figure*}

In order to represent words by vectors, word vectors were obtained from the
Word2vec model pre-trained on Wikipedia news by
FastText \url{https://fasttext.cc/docs/en/english-vectors.html}.  The vectors
had 300 dimensions. Since a robot is related to multiple words, the vectors were
averaged per dimension to form a single 300-dimension vector representing the
robot, as illustrated in Fig. \ref {fig:cCUq}.

In order to quantify the affect dimensions of words, we used the valence,
arousal and dominance (VAD) measurements in Mohammad's (2018) affect lexicon for
20,000 English word using best-worst scaling
\cite{hollis_scoring_2018,mohammad_obtaining_2018}.  After collecting, the
responses in the task were pre-processed to convert plural words into singular
forms and to replace capitalized words into lower-case words.  For compound
words and phrases, we primarily extracted the word that was more representative,
for example, ``canfly'' was replaced as ``fly'', ``car-like'' was replaced as
``car''; when representativeness was hard to decide, e.g. ``artificial
intelligence'', the whole compound was abandoned. Some adjectives were not in
the lexicon, but their synonyms were, e.g. ``gangly'' was not in the lexicon but
``awkward'' was in the lexicon. In this case, we replaced them with their
synonyms according to the Merriam-Webster thesaurus
(\url{https://www.merriam-webster.com/thesaurus/}). After preprocessing, the VAD
lexicon covered 92.33\% of the responses.  In the lexicon, the values of the
affect dimensions were ranged between 0 to 1, a larger value represents a higher
affect rating.

\section {RESULTS}

\begin{figure}[t]
  \centering
  \includegraphics[width=\linewidth] {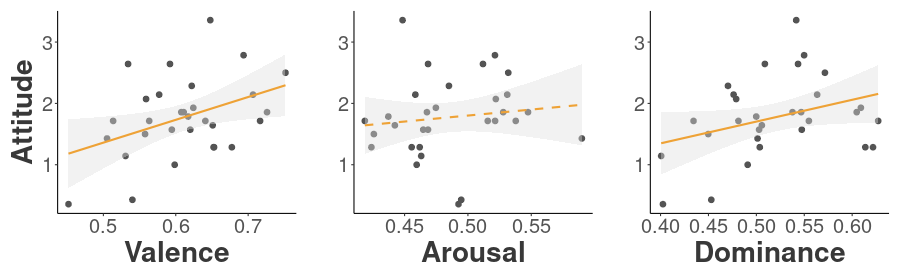}
  \caption {The relationships between the attitudes and dominance, valence,
    arousal. The dashed line indicates the insignificance in the corresponding
    LME analyses.}
  \label {fig:MhjS}
\end{figure}

\begin{figure}[t!]
  \centering
  \includegraphics[width=.9\linewidth] {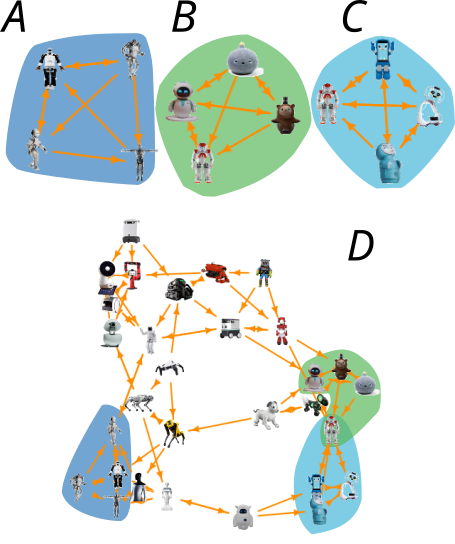}
  \caption{The network of robots based on their body image. (A-C) Three
    4-cliques in the network. (D) the whole network of all the 30 robots.
    The colors of the cliques correspond to the color in the network.
  }
  \label{fig:okO3}
\end{figure}

\begin{figure}[b!]
  \centering
  \includegraphics[width=.7\linewidth] {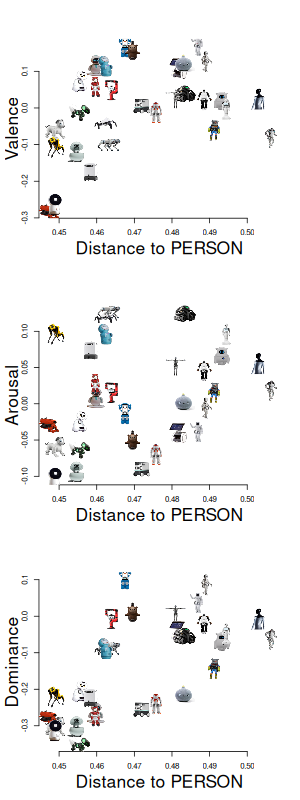}
  \caption{the relationship between affects and human distance. The x-axis is
    the distance between robot body images and the word ``person''; the y-axes
    are standardized valence, arousal and dominance.  }
  \label{fig:JLQD}
\end{figure}

\subsection {Word affects predicts subjective attitude}
At participant-level, we investigated whether the general attitude towards
robots can be predicted by affect measures of the association words.  Three
linear mixed-effect (LME) models were built with attitude scores as the
dependent variable and affect measures as the only fixed-effect independent
variable respectively. Robots with random intercepts were included as the
maximized random-effect structure.

The results showed that, valence is a strong predictor for attitude (likelihood
ratio test, LRT $\chi^2 = 15.577, p < .000$); dominance is a strong predictor
for attitude (LRT $\chi^2 = 19.529, p < .000$); whereas arousal didn't predict
attitude (LRT $\chi^2 = 2.101, p = .146$).

The relationships between the attitudes and affects were shown in Fig.
\ref{fig:MhjS}. The values of affects has been averaged across words within a
participant, i.e. each dot represents a participants. Overall, the results
validated that the semantic associations of a robot's body image reflect the
attitude of the observer.

\subsection {The graph of body images}
People usually classify robots in various approaches, such as the robot's shape,
the functions, and the usage, etc.  However, the relative positions between the
approaches, or which approach is the most preferred approach by subjects, have
scarcely been determined.

Here, we built a graph of robots based on the their association word vectors.
The vectors within a robot were averaged to represent the robot.  In the graph,
each robot connected to three of its most similar other robots.  based on the
vector representations and pair-wise cosine similarity, The connected robots
made up a graph of the 30 robots, as shown in Fig. \ref {fig:cCUq} and
Fig. \ref{fig:okO3}D.  In the body graph, we identified the some robot types
that are commonly referred to as humanoids, robot assistants, robot pets,
etc. However, some robots were not clearly classified as a member of a
category. For example, Aibo, who is often referred to as a ``robot dog'', is
closer connected to Eilik rather than to Spot; and Neubie, who is shaped as a
trolley or a locomotive, seems to be more likely to be connected with humanoid
rather than with similar-shaped Parky and other wheel-based ones, e.g. Pico and
Astro.

Even though, the most distinguishable categories are still identifiable by
finding out cliques, i.e.  the fully connected components in a graph.  We
underscored the 4-cliques in the graph to illustrate the robots categories, that
is (A) robot pets, (B) humanoids, and (C) robot kids (in Fig. \ref{fig:okO3}).
Nao seems to be perfectly fit into the categories of children-orient robots,
which include robot pets and robot kids. People who saw Nao would immediately
come up with the user base and usages into which it's supposed to fit.  In
contrast, Neubie, with its indefinite body image, is harder to be connected to a
specific user base or usage. Overall, the categories of robots identified by
their semantic relationships indicated that subjects tend to build body images
based on the expected user base and usage of the robot, but not based on their
shapes.

\subsection {Human distance predicts affects of body image}

The distance between a body image and a human-related word might be useful in
revealing how semantically close is the robot to human life. Here we obtained
the averaged vectors for robots, and computed their distance with the word
``person'' as $1-\text {similarity}_\text{cos} (v_1, v_2)$, where $v_1$ and
$v_2$ are the vectors abstracting a robot and the target word ``person''.

We chose ``person'' as the target word representing real human life out of the
reason to maximize the number of words near the distance between each robot and
the target word, which were used to control for the affect bias towards the
target word. An example of such a bias is the valence of words, which is
positive correlated with the words' distance to ``human''. Therefore, the affect
level of words tend to show a systemic relationship with the distance to the
target word, e.g. the distance to ``human''.  Therefore, we subtracted a
baseline from the value of affects of a robot to standardize the robot's affect
level.  The baseline is the mean affect level of words at a distance similar
with the robot.  Precisely, in the semantic space, we calculated the average
difference of the distances between robots and the target word, which is used as
a mask centering on the robot's distance to target word.  Masked words in the
affect lexicon were taken and used as affect baseline words. The baseline words
centering on both ``human'' and ``people'' were much fewer or fewer than those
centering on ``person''. Hence ``person'' as a target hold the most robust
baseline.

Here, we investigated the perceived affects by examining the relationship
between human distance and affects of body image.  Fig. \ref {fig:JLQD} shows
the results respectively for valence, arousal and dominance. We find that
valence and dominance are two functions of human distance. Valence grows with
human distance at first, and then start to plateau with a slight drop at a level
higher than the baseline; dominance grows monotonously with human distance. On
the other hand, no simple relationship between human distance and arousal is
identified.

Interestingly, we found most of the humanoids are further than others in human
distance. For example, the most distant robots are Atlas, Optimus, Pepper,
Musio, Hitchbot, Talos, iCub and Asimo; the nearest robots are Aquanaut, Jibo,
Spot, Aibo, Pico, Pyxel and Buddy. The results indicated that, for robots, being
more human-like might not be necessarily lead to fitting better into human life.

\section {DISCUSSION}

Psycholinguistics provides an enriched knowledge about words.  Here, we
introduce a paradigm that enables the utilization of this knowledge to analyze
human-robot interaction.

Specifically, we analyzed words associated with the visual image of robots to
explore humans' perception of them, and demonstrated that the word affect is
subject to people's attitudes towards robots.  Secondly, based on the associated
words, robots can be grouped according to their intended user base and
appearance.  The third result highlighted the importance of a robot's body image
and its implications for human-robot interaction.  The study concludes by
emphasizing the significance of body images in helping researchers and designers
understand the various dimensions of robots and the impact they have on human
responses.

Our paradigm consists of two procedures. In the first procedure, robots are
associated with a number of words, allowing us to infer perceived emotions,
i.e. the degree of valence (``positive/pleasure'' or ``negative/displeasure'')
and the degree of dominance (``powerful/strong'' or ``powerless/weak''). In
Fig. \ref {fig:MhjS}, we demonstrate how word-related valence and dominance can
be used to predict human attitudes towards robots from Likert scales. Free
association has also been used by another recent study on social robots
\cite{brondi_we_2021}. In that study, the authors investigated participants'
word associations with the written word ``robot'' and utilized factor analysis
to categorize the mental representation of robots as social entities. Our study
shares the same theoretical foundation with this previous research, but with a
more specific focus on individual robots rather than a general robot concept or
its social presence.

In the second procedure, the words are converted into vectors, abstracting a
robot with a 300-dimensional vector embedded in a robust semantic space. In
Fig. \ref {fig:okO3}, we demonstrate that although the grouping of the different robots is
based on the words that were associated with them, members of a group show
similarities of their physical appearance and derived features, e.g. humanoids
are grouped around the blue clique on the left side, next to the quadrupeds,
while child-oriented robots manifest themselves as the green and light blue
cliques. By increasing the number of robots (the present study only included 30
robots), this pattern might become even clearer. These results thus support the
notion that the body image of a robot encompasses interconnected visual,
semantic, and potentially social features, all as inferred from the robots'
morphology.

The paradigm provides an information-rich description of the robot perceived by
its human subject, which is expected to become increasingly stereoscopic by
probing the lexical scales and semantic space. For example, how these robots
relate to each other, and even to other concepts within the semantic space
(e.g. the concept of ``person''). In Fig. \ref {fig:JLQD}, we demonstrate how
valence and dominance relate to the proximity of the robot to the concept of
``person''.

Finally, the discovery of body image being predicted by valence and dominance
might also provide insights for commercial robot development. For example, in
the analysis shown in the right panel of Fig. \ref {fig:JLQD}, humanoids are
perceived as more dominant on the one hand, and further away from ``person'' on
the other hand. This suggests that the likelihood of purchasing a robot like
Jibo or Buddy (both low in dominance and close to ``person'') might be higher
than for humanoids.

\addtolength{\textheight}{-14.3cm} %

\bibliographystyle{IEEEtran}
\bibliography{arso2023}

\end{document}